\documentclass[conference]{IEEEtran}

\makeatletter

\def\ps@IEEEtitlepagestyle{%
  \def\@evenfoot{}%
}

\usepackage{eso-pic}
\IEEEoverridecommandlockouts
\usepackage{cite}
\usepackage{amsmath,amssymb,amsfonts}
\usepackage{algorithmic}
\usepackage{graphicx}
\usepackage{textcomp}
\usepackage{xcolor}
\def\BibTeX{{\rm B\kern-.05em{\sc i\kern-.025em b}\kern-.08em
    T\kern-.1667em\lower.7ex\hbox{E}\kern-.125emX}}

\usepackage{eso-pic}
\newcommand\AtPageUpperMyright[1]{\AtPageUpperLeft{%
 \put(\LenToUnit{0.17\paperwidth},\LenToUnit{-2cm}){%
     \parbox{0.9\textwidth}{\raggedleft\fontsize{8}{11}\selectfont #1}}%
 }}%
\newcommand{\conf}[1]{%
\AddToShipoutPictureBG*{%
\AtPageUpperMyright{#1}
}
}

\begin{document}
\title{\vspace*{1cm} Comparing Chunking and Embedding Strategies for Turkish RAG Systems\\
}

\author{\IEEEauthorblockN{Mustafa Sertaç Türkel}
\IEEEauthorblockA{\textit{Data Science and Innovation} \\
\textit{Ata Technology Platforms}\\
Istanbul, Turkey \\
sertac.turkel@atptech.com}
\and
\IEEEauthorblockN{Fatma Nur Korkmaz}
\IEEEauthorblockA{\textit{Data Science and Innovation} \\
\textit{Ata Technology Platforms}\\
Istanbul, Turkey \\
fatmanur.korkmaz@atptech.com}
\and
\IEEEauthorblockN{Ahmet Tuğrul Bayrak}
\IEEEauthorblockA{\textit{Data Science and Innovation} \\
\textit{Ata Technology Platforms}\\
Istanbul, Turkey \\
tugrul.bayrak@atptech.com}
}

\maketitle
\conf{\textit{ 6. Interdisciplinary Conference on Electrics and Computer (INTCEC 2026) \\
24-25 September 2026, Chicago-USA}}

\begin{center}
\small
Accepted to INTCEC 2026. This is the author's pre-print version. The final authenticated version will be available through the conference proceedings.
\end{center}

\begin{abstract}
Retrieval-Augmented Generation conditions a language model on chunks retrieved from a document collection. Its accuracy is therefore limited by the chunking and embedding stages that determine what can be retrieved. We compare Turkish document question answering across three chunking strategies (fixed-length, semantic, and layout-aware Docling), five embedding models, and two LLMs, over three documents with contrasting layouts. Every configuration answers the same question set, which allows component effects to be separated by paired testing rather than inferred from separate benchmarks. The fully crossed design yields 9{,}000 graded question-answer evaluations, each scored by an independent judge model, and component comparisons are tested by paired McNemar tests under Holm correction. The three leading embedding models are statistically indistinguishable, so language specialization yields no measurable retrieval advantage. The faster LLM is not the more accurate one. The preferred configuration depends on content type, since layout-aware chunking helps table-heavy documents far more than text-heavy ones.

\end{abstract}


\section{Introduction}

Retrieval-Augmented Generation (RAG) combines a retrieval component that selects relevant passages from a document collection with a generative large language model (LLM) that answers conditioned on those passages \cite{lewis2020rag, izacard2021fid}. Because the LLM answers from retrieved context, quality depends on decisions made before generation: how documents are split into retrievable units (\textit{chunking}), and how those units and the query are mapped into a vector space for similarity search (\textit{embedding}) \cite{karpukhin2020dpr}.  These choices interact with each other and with document properties, yet in practice are often made by convention rather than measurement \cite{gao2023ragsurvey}.

These choices matter particularly for Turkish. As an agglutinative language, Turkish encodes grammatical information through extensive suffixation, so a single lemma appears in many inflected forms \cite{schweter2020berturk}. The effect manifests first at tokenization: subword vocabularies poorly matched to Turkish morphology fragment inflected forms inconsistently, degrading downstream Turkish model performance \cite{toraman2023tokenization}, and morphological sparsity is a recurring obstacle for morphologically rich languages \cite{sahin2018morphing}. An embedding model inheriting such a mismatch may place semantically similar passages far apart. Document layout adds a second factor: institutional documents mix text with tables, and a chunking method that splits a table across boundaries makes the corresponding facts harder to retrieve regardless of the embedding.

This paper measures, by direct comparison, the components of an effective RAG configuration for Turkish document question answering. We treat this as a controlled comparison across chunking strategy, embedding model, and LLM, evaluating every combination on the same question set. Operational metrics are reported alongside accuracy, because a deployed system must balance quality against cost and latency. Our contributions are threefold. First, we describe a reproducible harness that grades 9{,}000 question-answer pairs across a fully crossed design, reporting all fourteen planned component comparisons with a paired significance test under Holm correction. Second, we show that the three leading embedding models cannot be statistically separated, and that language specialization does not by itself confer a retrieval advantage. Third, we show that the preferred configuration depends on content type.

\section{Related Work}

RAG combines the parametric knowledge of a language model with non-parametric retrieval over an external corpus, allowing use of up-to-date or domain-specific content without retraining \cite{lewis2020rag}. Generative readers such as Fusion-in-Decoder showed that conditioning generation on several retrieved passages jointly improves open-domain question answering \cite{izacard2021fid}. Retrieval has shifted from sparse lexical matching to dense retrieval, in which queries and passages are embedded into a shared space and compared by inner product or cosine similarity \cite{karpukhin2020dpr}; at scale this uses approximate nearest-neighbour indexes such as FAISS \cite{johnson2021faiss}. Embedding quality therefore bounds end-to-end retrieval performance.

Sentence-level encoders trained with Siamese or contrastive objectives, such as Sentence-BERT \cite{reimers2019sbert} and language-agnostic models such as LaBSE \cite{feng2022labse}, suit semantic similarity better than token-level masked language models \cite{devlin2019bert} or classical subword word embeddings \cite{bojanowski2017fasttext}. Large multilingual encoders such as multilingual-e5 \cite{wang2024multilinguale5} and commercial embedding services \cite{openai2024embeddings} extend dense representations to many languages at once, but strong average multilingual performance does not guarantee strong results for any single language. For Turkish, dedicated pretrained models such as BERTurk \cite{schweter2020berturk} outperform multilingual models on downstream tasks such as classification and offensive language detection \cite{safaya2020kuisail}, and tokenization studies show that the match between subword vocabulary and Turkish morphology measurably affects model quality \cite{toraman2023tokenization}. Whether this transfers from classification to retrieval, where the objective is relative ranking rather than label prediction, is an open question we examine directly.

Chunking has received less systematic attention, though it determines which units the retriever can see at all. The problem predates RAG as \textit{text segmentation}: lexical-cohesion methods such as TextTiling place boundaries where vocabulary overlap between adjacent windows drops \cite{hearst1997texttiling}, and supervised neural segmenters learn boundary placement from labelled documents \cite{koshorek2018segmentation}. Fixed-length splitting remains the common baseline in deployed systems, semantic segmentation is the practical descendant of the cohesion-based line, and layout-aware parsing preserves structural elements such as tables and headings, the role played by toolkits such as Docling \cite{docling2024}. For evaluation, model-based protocols such as RAGAS assess RAG outputs at scale \cite{es2024ragas}, and surveys review the design space \cite{gao2023ragsurvey}. We contribute a controlled, fully crossed comparison for Turkish documents mixing textual and tabular content, with operational metrics and significance tests.

\section{Data}

The corpus consists of three institutional Turkish documents with contrasting layout characteristics, with a fixed question set per document whose reference answers are known in advance. The documents are distinguished by the proportion of content presented as tables rather than text and are referred to by document type: \textit{table-heavy}, \textit{balanced} (tables and text in comparable proportion), and \textit{text-heavy}. Identical chunking and embedding methods are applied to all three, holding those factors constant across layout types.

Each document has 100 distinct questions, 300 in total, and every question is labelled according to whether answering it requires information located in a table, which supports the separate analysis in Section V. Of the 300 questions, 105 are table-grounded and 195 text-grounded, distributed as 80, 22, and 3 of 100 in the table-heavy, balanced, and text-heavy documents respectively.

Crossing 3 chunking strategies, 5 embedding models, and 2 LLMs gives 30 pipeline configurations, and answering all 300 questions under each yields $300\times30=9{,}000$ graded evaluations. Each configuration-level accuracy is a mean over 300 questions, each embedding-level figure a mean over 1{,}800 evaluations, each chunking-level figure a mean over 3{,}000, and each LLM-level figure a mean over 4{,}500. Because the design is fully crossed, component comparisons are not confounded by uneven allocation of questions or documents. Since the same 300 questions underlie every configuration, the paired tests of Section IV-F apply.

\section{Method}

\subsection{RAG Pipeline}
\begin{figure}[!t]
\centering
\includegraphics[width=\columnwidth]{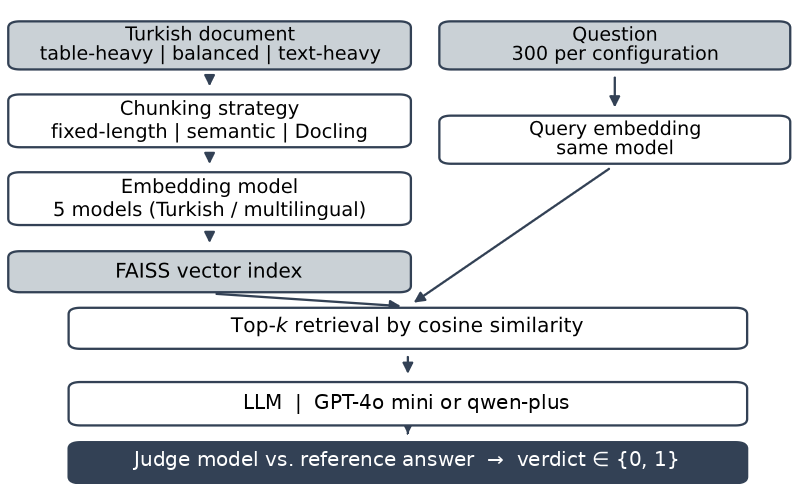}
\vspace{0.4mm}\hrule\vspace{0.6mm}
{\scriptsize
\begin{algorithmic}[1]
\STATE index $\leftarrow$ FAISS(embedding model(chunker(document)))
\FOR{each question}
  \STATE answer $\leftarrow$ LLM(prompt(question, top-$k$(index, question)))
  \STATE verdict $\leftarrow$ judge(question, answer, reference) $\in \{0,1\}$
\ENDFOR
\end{algorithmic}
}
\vspace{0.4mm}\hrule
\caption{Architecture and procedure of one RAG configuration}
\label{fig:pipeline}
\end{figure}

Every configuration follows the same pipeline, differing only in the components under comparison. Each document is segmented by the selected chunking strategy; every chunk is encoded into a dense vector by the selected embedding model and stored in a FAISS index \cite{johnson2021faiss}. At query time the question is embedded with the same model, the most similar chunks are retrieved by cosine similarity, and they are inserted into a fixed prompt template with the question. The selected LLM produces the answer, which is graded against the reference as described in Section IV-E. Fig.~\ref{fig:pipeline} shows the architecture of a single configuration together with the corresponding procedure. Because the number of retrieved chunks, the prompt template, and the similarity metric were held constant, the differences in Section V are attributable to the three factors under study rather than to incidental changes in retrieval or prompting.

\subsection{Chunking Strategies}
\textit{Fixed-length} chunking splits text into chunks of fixed target character length regardless of content, serving as a baseline. \textit{Semantic} chunking places boundaries where topical similarity between adjacent sentences drops; it is the practical descendant of lexical-cohesion segmentation \cite{hearst1997texttiling, koshorek2018segmentation} and typically produces shorter chunks. \textit{Docling} chunking is layout-aware, parsing document structure and keeping structurally coherent elements, in particular tables, together within a chunk \cite{docling2024}, which yields larger chunks. The three strategies thus represent increasing awareness of structure, from none, through topical structure, to explicit layout.

\subsection{Embedding Models}
\textit{FastText} \cite{bojanowski2017fasttext} is a classical subword-based word-embedding method included as a lightweight baseline. \textit{TurkEmbed} \cite{ezerceli2025turkembed} and \textit{Mursit-large} \cite{ugur2026mecellem} are publicly available embedding models oriented toward Turkish, the former trained on Turkish natural language inference and sentence-similarity data and the latter retrieval-tuned on a Turkish-dominant pretraining corpus. \textit{multilingual-e5-large} \cite{wang2024multilinguale5} is a large multilingual text-embedding model, and \textit{text-embedding-3-small} \cite{openai2024embeddings} a general-purpose commercial model accessed through a hosted API. All five were evaluated in the identical pipeline, with only the embedding step varying. Dense sentence encoders of this kind \cite{reimers2019sbert, feng2022labse} have largely replaced word-level representations \cite{devlin2019bert} for retrieval; the selection lets us test whether that holds for Turkish and whether language-specialized models add further benefit.

\subsection{LLMs}
Two LLMs were compared: \textit{GPT-4o mini} \cite{openai2024gpt4omini} and \textit{qwen-plus} \cite{alibaba2024qwenplus}, compact commercial models from different providers. Both received the same template and retrieved context, which isolates generation quality and latency from retrieval.

\subsection{Evaluation Protocol and Operational Metrics}
Each answer was graded correct or incorrect against the reference for its question, and accuracy is the percentage correct over the relevant set. Grading was automatic, performed by a judge model, Llama 3.3 70B Instruct \cite{meta2024llama33}, prompted with the question, the reference answer, and the candidate answer and instructed to return a binary verdict. This reference-guided model-based protocol follows practice established for scalable RAG evaluation \cite{es2024ragas}. The judge model is deliberately drawn from a third provider, distinct from both LLMs under comparison, which prevents grading from favouring either through self-preference.

Three operational metrics were recorded per evaluation: \textit{average chunk size}, the mean character length of retrieved chunks and a proxy for context tokens and therefore cost per query; \textit{average answer length}, the mean character length of generated answers; and \textit{average response time}, mean end-to-end API latency.

\subsection{Statistical Analysis}
Component comparisons exploit the fact that all configurations answer the same 300 questions: for two levels of a factor the evaluations match one-to-one on the remaining factors. Differences are therefore assessed with the paired McNemar test with continuity correction, rather than a test for independent proportions, which would ignore the pairing. With fourteen comparisons planned, $p$-values are adjusted by Holm-Bonferroni and the adjusted values interpreted.

\section{Results}

Across all 9{,}000 evaluations, overall accuracy is 68.10\%, and it varies widely with configuration.

Table \ref{tab:embedding} reports accuracy and operational metrics by embedding model, each over 1{,}800 evaluations. multilingual-e5-large is strongest at 80.00\%, ahead of the Turkish-oriented models and text-embedding-3-small, which fall between 75.78\% and 78.67\%; FastText is substantially lower at 27.56\%, which rules out a subword word-embedding method in this setting. The four competitive models differ only modestly on operational metrics: chunk sizes range from 1{,}836 to 1{,}909 characters and response times from 3.99 to 4.10 s. Their accuracy differences are therefore not explained by context length.

\begin{table}[htbp]
\caption{Performance by Embedding Model}
\begin{center}
\scriptsize
\setlength{\tabcolsep}{3pt}
\begin{tabular}{|l|c|c|c|c|}
\hline
\textbf{Embedding} & \textbf{Acc. (\%)} & \textbf{Chunk} & \textbf{Ans.} & \textbf{Time} \\
 & & \textbf{size} & \textbf{len.} & \textbf{(s)} \\ \hline
multilingual-e5-large & 80.00 & 1836.39 & 72.63 & 3.99 \\ \hline
TurkEmbed & 78.67 & 1865.79 & 73.34 & 4.04 \\ \hline
text-embedding-3-small & 78.50 & 1869.44 & 77.54 & 4.10 \\ \hline
Mursit-large & 75.78 & 1909.23 & 74.54 & 4.09 \\ \hline
FastText & 27.56 & 1960.40 & 86.28 & 4.33 \\ \hline
\end{tabular}
\label{tab:embedding}
\end{center}
\end{table}

Between the two LLMs, GPT-4o mini is more accurate than qwen-plus (69.58\% versus 66.62\%), while qwen-plus is faster, averaging 2.60 s against 5.62 s; mean answer lengths are 79.55 and 74.18 characters. The accuracy gap is small in absolute terms but statistically reliable (Table \ref{tab:significance}). This is a cost-quality trade-off: qwen-plus responds in under half the time at a cost of 2.96 points of accuracy, acceptable in latency-sensitive deployments and unsuitable where accuracy is the binding constraint.

\begin{figure}[!t]
\centering
\includegraphics[width=0.82\columnwidth]{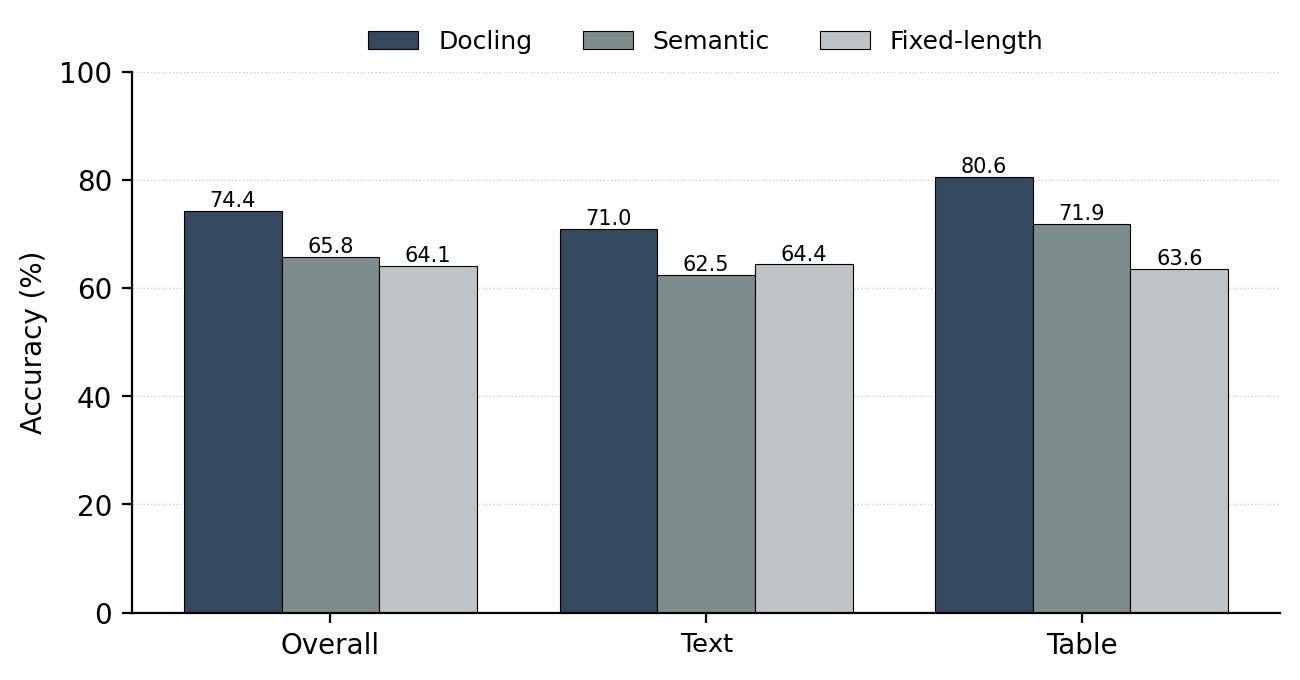}
\caption{Accuracy by chunking strategy, overall and by question type}
\label{fig:chunking}
\end{figure}

Fig.~\ref{fig:chunking} reports accuracy by chunking strategy, decomposed by question type. Docling is strongest overall at 74.37\%, and its advantage is largest on table-grounded questions, consistent with keeping tables intact. Semantic and fixed-length chunking are close overall, but their relative order reverses by question type: semantic is better on table questions , fixed-length slightly better on text questions. Docling achieves this with chunks averaging 3{,}195 characters, 2.2 times the 1{,}451 of fixed-length and the 1{,}019 of semantic chunking. The larger chunks raise per-query context cost, which is the principal cost of this advantage. Mean response times are close across the three strategies, between 3.72 and 4.34 s.

\subsection{Significance of Component Differences}

Table \ref{tab:significance} reports the paired McNemar comparisons of Section IV-F with Holm-adjusted $p$-values. Ten of the fourteen planned comparisons remain significant, including all four comparisons of the dense embedding models against FastText. The four that are not significant are informative: multilingual-e5-large cannot be separated from TurkEmbed or text-embedding-3-small, TurkEmbed and text-embedding-3-small are indistinguishable, and semantic and fixed-length chunking cannot be separated. In contrast, both comparisons of Docling against the alternatives, the LLM comparison, and all three comparisons against Mursit-large are significant. The chunking and LLM effects are statistically reliable.

\begin{table}[htbp]
\centering
\scriptsize
\setlength{\tabcolsep}{3pt}
\caption{Paired McNemar Comparisons with Holm Correction}
\begin{tabular}{|l|c|c|c|}
\hline
\textbf{Comparison} & \textbf{Diff. (pts)} & \textbf{$p$} & \textbf{$p_{\text{Holm}}$} \\ \hline
Docling vs.\ Fixed-length & $+10.23$ & $8{\times}10^{-29}$ & $<0.001$ \\ \hline
Docling vs.\ Semantic & $+8.57$ & $1{\times}10^{-23}$ & $<0.001$ \\ \hline
GPT-4o mini vs.\ qwen-plus & $+2.96$ & $7{\times}10^{-11}$ & $<0.001$ \\ \hline
mE5-large vs.\ Mursit-lg & $+4.22$ & $1.5{\times}10^{-5}$ & $0.0001$ \\ \hline
TurkEmbed vs.\ Mursit-lg & $+2.89$ & $0.0034$ & $0.021$ \\ \hline
te3-small vs.\ Mursit-lg & $+2.72$ & $0.0065$ & $0.033$ \\ \hline
Semantic vs.\ Fixed-length & $+1.67$ & $0.053$ & $0.211$ \\ \hline
mE5-large vs.\ te3-small & $+1.50$ & $0.105$ & $0.315$ \\ \hline
mE5-large vs.\ TurkEmbed & $+1.33$ & $0.162$ & $0.323$ \\ \hline
TurkEmbed vs.\ te3-small & $+0.17$ & $0.900$ & $0.900$ \\ \hline
\end{tabular}

\label{tab:significance}
\end{table}

\subsection{Accuracy by Chunking and Embedding Model}

The fully crossed design also allows the two retrieval-side factors to be examined jointly. Table \ref{tab:interaction} reports accuracy for each of the fifteen chunking-by-embedding pairs, each over 600 evaluations. Under Docling chunking the four modern embedding models fall within 1.33 points of one another, a negligible spread. Under semantic and fixed-length chunking the same four span 7.67 and 7.17 points, and the best embedding model changes: Mursit-large under Docling, multilingual-e5-large under semantic, and TurkEmbed under fixed-length. Along the other axis, sensitivity to the chunking strategy ranges from 7.00 points for TurkEmbed to 15.00 for Mursit-large. FastText varies by 22.17 points.

\begin{table}[htbp]
\centering
\scriptsize
\setlength{\tabcolsep}{3pt}
\caption{Accuracy (\%) by Chunking Strategy and Embedding Model}
\begin{tabular}{|l|c|c|c|c|c|}
\hline
\textbf{Chunking} & \textbf{FastText} & \textbf{mE5-large} & \textbf{Mursit-lg} & \textbf{te3-small} & \textbf{TurkEmbed} \\ \hline
Docling & 39.50 & 82.50 & 83.83 & 83.00 & 83.00 \\ \hline
Semantic & 17.33 & 82.33 & 74.67 & 77.67 & 77.00 \\ \hline
Fixed-length & 25.83 & 75.17 & 68.83 & 74.83 & 76.00 \\ \hline
\end{tabular}

\label{tab:interaction}
\end{table}

Under Docling a strong embedding model can therefore be chosen on cost, licensing, or latency grounds with little penalty, whereas under the two strategies that ignore layout the choice among the same embedding models spans up to 7.17 points, and the weakest of them, Mursit-large, falls 15.00 points below its own best case. This also qualifies the ranking in Table \ref{tab:embedding}, which averages over chunking strategies and is not stable within them.

\subsection{Performance by Document Type}

Table \ref{tab:doctype} reports accuracy by document type. The balanced document is markedly the most difficult, 57.73\% against 74.57\% and 72.00\%. The benefit of layout-aware chunking is uneven. On the text-heavy document the three strategies lie within 2.1 points, leaving chunking little influence there. On the table-heavy and balanced documents Docling leads the best alternative by 6.0 and 12.3 points. In the table-heavy document, table questions trail text questions only because Table \ref{tab:doctype} averages across chunking strategies. Fixed-length chunking scores 61.50\% on those questions against 79.50\% on text questions, whereas under Docling table questions lead (82.62 versus 81.00).

\begin{table}[htbp]
\centering
\footnotesize
\setlength{\tabcolsep}{4pt}
\caption{Accuracy (\%) by Document Type}
\begin{tabular}{|l|c|c|c|}
\hline
\textbf{Document type} & \textbf{Overall (\%)} & \textbf{Text Q.} & \textbf{Table Q.} \\ \hline
Table-heavy & 74.57 & 78.50 & 73.58 \\ \hline
Balanced & 57.73 & 54.49 & 69.24 \\ \hline
Text-heavy & 72.00 & 72.65 & 51.11 \\ \hline
\end{tabular}

\label{tab:doctype}
\end{table}

\subsection{Best Complete Configurations}

Table \ref{tab:top} reports the five best configurations overall and on table-grounded questions. The best overall, Docling chunking with Mursit-large and GPT-4o mini, reaches 87.00\%, 18.90 points above the 68.10\% average. Docling appears in eight of the ten best overall configurations and GPT-4o mini in five. The five configurations ranked sixth to tenth all use qwen-plus, spanning 80.67-81.33\% at 2.04-3.63 s, and remain within 6.5 points of the best configuration at roughly half its latency. The top of the ranking is consistent in its chunking strategy while varying in embedding model and LLM.

The subset rankings differ: the configuration best on average is not best on every kind of question. On table questions the leaders reach 93.33\% and are dominated by Docling and semantic chunking, and a qwen-plus configuration reaches 91.43\% at 2.28 s, narrowing the latency-accuracy gap. On text questions accuracy is lower, the best configuration reaching 84.10\%. A fixed-length configuration enters the ten best there while none does so on table questions. This is consistent with fixed-length chunking being less damaging for text than for tables, where an ill-placed boundary can separate a value from its row or column header.

\begin{table}[htbp]
\centering
\scriptsize
\setlength{\tabcolsep}{2pt}
\caption{Top Five Configurations Overall and on Table-Grounded Questions}
\begin{tabular}{|l|l|l|c|c|}
\hline
\textbf{Chunking} & \textbf{Embedding} & \textbf{LLM} & \textbf{Acc. (\%)} & \textbf{Time (s)} \\ \hline
\multicolumn{5}{|l|}{\textit{Overall}} \\ \hline
Docling & Mursit-lg & GPT-4o mini & 87.00 & 5.17 \\ \hline
Docling & te3-small & GPT-4o mini & 85.00 & 4.98 \\ \hline
Docling & TurkEmbed & GPT-4o mini & 84.67 & 4.86 \\ \hline
Docling & mE5-large & GPT-4o mini & 84.00 & 4.91 \\ \hline
Semantic & mE5-large & GPT-4o mini & 83.33 & 5.14 \\ \hline
\multicolumn{5}{|l|}{\textit{Table-related questions}} \\ \hline
Docling & TurkEmbed & GPT-4o mini & 93.33 & 4.29 \\ \hline
Docling & mE5-large & GPT-4o mini & 93.33 & 4.26 \\ \hline
Docling & Mursit-lg & GPT-4o mini & 92.38 & 4.50 \\ \hline
Semantic & mE5-large & GPT-4o mini & 92.38 & 4.61 \\ \hline
Semantic & te3-small & qwen-plus & 91.43 & 2.28 \\ \hline
\end{tabular}

\label{tab:top}
\end{table}

\section{Discussion}

First, the chunking strategy is the most influential factor for content containing tabular structure. Docling chunking improves table-question accuracy by 17.0 points over the fixed-length baseline and by 8.7 points over semantic chunking, because it retrieves whole tables rather than fragments. Both comparisons survive correction for multiple testing. The gain comes with larger chunks and therefore higher context-token consumption, and it is conditional: on the text-heavy document the three strategies differ by at most 2.1 points, leaving predominantly textual content little to gain from layout-aware chunking.

Second, the effect of embedding quality is concentrated among weak embedding models. The four modern embedding models differ by 4.22 points overall and the top three cannot be statistically separated, whereas FastText is substantially lower at 27.56\%; the important decision is to avoid an inadequate embedding model rather than to choose among the strong ones. Table \ref{tab:interaction} adds that under Docling the four span 1.33 points, while under fixed-length chunking they span 7.17 points and Mursit-large falls 15.00 points below its own best case.

\subsection{Language Specialization and Retrieval}

We expected Turkish morphology to favour Turkish-oriented embedding models. The results instead show equivalence. multilingual-e5-large is nominally most accurate but not significantly ahead of TurkEmbed or text-embedding-3-small, which are themselves indistinguishable. \textit{Language specialization confers no measurable retrieval advantage on this corpus}: a strong multilingual embedding model, a Turkish-oriented one, and a commercial multilingual service perform equivalently, and only Mursit-large is reliably behind. For practitioners the implication is direct: the embedding model can be chosen on cost, latency, or licensing grounds without an accuracy penalty.

The morphological motivation is nonetheless supported, but at a different level of the pipeline than anticipated. FastText, a word-level model built from subword $n$-grams without contextual composition, falls to 27.56\% and is separated from every dense embedding model by 48.22 to 52.44 points at $p_{\text{Holm}}<0.001$. This is consistent with the failure mode agglutinative morphology would predict, since representations that do not compose morphology contextually are less able to bring inflected variants of a lemma together. Because the design contains no non-agglutinative control language, however, this gap cannot be attributed to Turkish morphology specifically rather than to the absence of contextual composition in general. What large multilingual embedding models appear to provide is sufficient subword coverage and contextual composition to absorb Turkish morphology without language-specific pretraining, consistent with tokenization work locating the effect in the match between subword vocabulary and morphology rather than in the language identity of the training corpus \cite{toraman2023tokenization}. Transfer from classification, where Turkish-specific pretraining does help \cite{schweter2020berturk, safaya2020kuisail}, to retrieval is thus not automatic.

\subsection{LLM Choice and Trade-offs}

The faster LLM is not the more accurate one. qwen-plus answers in under half the time of GPT-4o mini while giving up 2.96 points, small but statistically reliable ($p_{\text{Holm}}<0.001$), and a qwen-plus configuration ranks among the five best on table questions. The LLM choice therefore depends on whether a deployment prioritizes accuracy or latency, and can be made largely independently of chunking and embedding, whose leading settings are stable across both LLMs. For Turkish document question answering over mixed content, these results favour layout-aware chunking with any of the three leading embedding models, reserving the faster LLM for latency-sensitive settings and the more accurate one where correctness dominates.

\section{Conclusion}

We compared chunking and embedding strategies for Turkish-language Retrieval-Augmented Generation over a fully crossed design of three chunking methods, five embedding models, and two LLMs, yielding 9{,}000 graded question-answer pairs from documents of three contrasting layout types. Docling chunking, multilingual-e5-large, and GPT-4o mini were the best individual components, yet they did not compose into the best complete configuration: the highest-scoring configuration paired Docling with Mursit-large and GPT-4o mini at 87.00\%, 18.90 points above the average, even though Mursit-large ranked last of the four modern embedding models in the marginal comparison. Under Docling chunking the four modern embedding models fall within 1.33 points of one another, so the marginal ranking is not stable within chunking strategies. Paired tests show which differences are reliable: the chunking and LLM effects are, whereas the ranking of the three leading embedding models is not, from which we conclude that language specialization confers no measurable retrieval advantage on this corpus. The preferred configuration also depends on content type, and accuracy should be weighed against cost and latency. Future work includes a larger and more varied corpus, validation of the automatic grades against human annotation on a stratified sample, and hybrid chunking that routes tabular and textual content to different strategies within one document.


\begin{thebibliography}{00}

\bibitem{lewis2020rag} P. Lewis et al., ``Retrieval-Augmented Generation for Knowledge-Intensive NLP Tasks,'' in \textit{Proc. Adv. Neural Inf. Process. Syst. (NeurIPS)}, vol. 33, 2020, pp. 9459--9474.

\bibitem{izacard2021fid} G. Izacard and E. Grave, ``Leveraging Passage Retrieval with Generative Models for Open Domain Question Answering,'' in \textit{Proc. Conf. Eur. Chapter Assoc. Comput. Linguist. (EACL)}, 2021, pp. 874--880.

\bibitem{karpukhin2020dpr} V. Karpukhin et al., ``Dense Passage Retrieval for Open-Domain Question Answering,'' in \textit{Proc. Conf. Empirical Methods Natural Lang. Process. (EMNLP)}, 2020, pp. 6769--6781.

\bibitem{gao2023ragsurvey} Y. Gao et al., ``Retrieval-Augmented Generation for Large Language Models: A Survey,'' arXiv preprint arXiv:2312.10997, 2023.

\bibitem{schweter2020berturk} S. Schweter, ``BERTurk: BERT Models for Turkish,'' Zenodo, 2020, doi: 10.5281/zenodo.3770924.

\bibitem{toraman2023tokenization} C. Toraman, E. H. Yilmaz, F. Şahinuç, and O. Ozcelik, ``Impact of Tokenization on Language Models: An Analysis for Turkish,'' \textit{ACM Trans. Asian Low-Resour. Lang. Inf. Process.}, vol. 22, no. 4, pp. 1--21, 2023.

\bibitem{sahin2018morphing} G. G. Şahin and M. Steedman, ``Data Augmentation via Dependency Tree Morphing on Low-Resource Languages,'' in \textit{Proc. Conf. Empirical Methods Natural Lang. Process. (EMNLP)}, 2018, pp. 5004--5009.

\bibitem{johnson2021faiss} J. Johnson, M. Douze, and H. Jégou, ``Billion-Scale Similarity Search with GPUs,'' \textit{IEEE Trans. Big Data}, vol. 7, no. 3, pp. 535--547, 2021.

\bibitem{reimers2019sbert} N. Reimers and I. Gurevych, ``Sentence-BERT: Sentence Embeddings using Siamese BERT-Networks,'' in \textit{Proc. Conf. Empirical Methods Natural Lang. Process. (EMNLP)}, 2019, pp. 3982--3992.

\bibitem{feng2022labse} F. Feng, Y. Yang, D. Cer, N. Arivazhagan, and W. Wang, ``Language-agnostic BERT Sentence Embedding,'' in \textit{Proc. Annu. Meeting Assoc. Comput. Linguist. (ACL)}, 2022, pp. 878--891.

\bibitem{devlin2019bert} J. Devlin, M.-W. Chang, K. Lee, and K. Toutanova, ``BERT: Pre-training of Deep Bidirectional Transformers for Language Understanding,'' in \textit{Proc. Conf. North Amer. Chapter Assoc. Comput. Linguist.: Human Lang. Technol. (NAACL-HLT)}, 2019, pp. 4171--4186.

\bibitem{bojanowski2017fasttext} P. Bojanowski, E. Grave, A. Joulin, and T. Mikolov, ``Enriching Word Vectors with Subword Information,'' \textit{Trans. Assoc. Comput. Linguist.}, vol. 5, pp. 135--146, 2017.

\bibitem{wang2024multilinguale5} L. Wang et al., ``Multilingual E5 Text Embeddings: A Technical Report,'' arXiv preprint arXiv:2402.05672, 2024.

\bibitem{openai2024embeddings} OpenAI, ``Embeddings,'' OpenAI API Documentation, 2024. [Online]. Available: https://platform.openai.com/docs/guides/embeddings

\bibitem{safaya2020kuisail} A. Safaya, M. Abdullatif, and D. Yuret, ``KUISAIL at SemEval-2020 Task 12: BERT-CNN for Offensive Speech Identification in Social Media,'' in \textit{Proc. 14th Workshop Semantic Eval. (SemEval)}, 2020, pp. 2054--2059.

\bibitem{hearst1997texttiling} M. A. Hearst, ``TextTiling: Segmenting Text into Multi-paragraph Subtopic Passages,'' \textit{Comput. Linguist.}, vol. 23, no. 1, pp. 33--64, 1997.

\bibitem{koshorek2018segmentation} O. Koshorek, A. Cohen, N. Mor, M. Rotman, and J. Berant, ``Text Segmentation as a Supervised Learning Task,'' in \textit{Proc. Conf. North Amer. Chapter Assoc. Comput. Linguist.: Human Lang. Technol. (NAACL-HLT)}, 2018, pp. 469--473.

\bibitem{docling2024} C. Auer et al., ``Docling Technical Report,'' arXiv preprint arXiv:2408.09869, 2024.

\bibitem{es2024ragas} S. Es, J. James, L. Espinosa-Anke, and S. Schockaert, ``RAGAS: Automated Evaluation of Retrieval Augmented Generation,'' in \textit{Proc. Conf. Eur. Chapter Assoc. Comput. Linguist.: Syst. Demonstrations (EACL)}, 2024, pp. 150--158.

\bibitem{ezerceli2025turkembed} Ö. Ezerceli, G. Gümüşçekiçci, T. Erkoç, and B. Özenç, ``TurkEmbed: Turkish Embedding Model on Natural Language Inference and Sentence Text Similarity Tasks,'' in \textit{Proc. IEEE 11th Int. Conf. Adv. Softw., Hardware Syst. Eng. (ASYU)}, 2025, doi: 10.1109/ASYU67174.2025.11208511.

\bibitem{ugur2026mecellem} Ö. Uğur et al., ``Mecellem Models: Turkish Models Trained from Scratch and Continually Pre-trained for the Legal Domain,'' arXiv preprint arXiv:2601.16018, 2026.

\bibitem{openai2024gpt4omini} OpenAI, ``GPT-4o mini,'' OpenAI Model Documentation, 2024. [Online]. Available: https://platform.openai.com/docs/models

\bibitem{alibaba2024qwenplus} Alibaba Cloud, ``Qwen-Plus,'' Alibaba Cloud Model Studio Documentation, 2024. [Online]. Available: https://www.alibabacloud.com/help/en/model-studio

\bibitem{meta2024llama33} Meta AI, ``Llama 3.3 70B Instruct,'' Model card, 2024. [Online]. Available: https://huggingface.co/meta-llama/Llama-3.3-70B-Instruct

\end{thebibliography}
\end{document}